\documentclass[letterpaper, 10 pt, conference]{ieeeconf}  

\IEEEoverridecommandlockouts 

\usepackage{graphicx}
\usepackage{textgreek}
\usepackage{comment}
\usepackage{stfloats}
\usepackage{multirow}

\title{\LARGE \bf
UltraBot: Autonomous Mobile Robot for Indoor UV-C Disinfection
}

\author{Stepan Perminov, Nikita Mikhailovskiy, Alexander Sedunin, Iaroslav Okunevich,\\
Ivan Kalinov, Mikhail Kurenkov, and Dzmitry Tsetserukou
\thanks{All authors are with the Intelligent Space Robotics Laboratory, Space CREI, Skolkovo Institute of Science and Technology, Moscow, Russian Federation.
        {\tt \{stepan.perminov, nikita.mikhailovskiy, alexander.sedunin, iaroslav.okunevich, i.kalinov, mikhail.kurenkov, d.tsetserukou\}@skoltech.ru}}%

}

\begin{document}

\maketitle
\thispagestyle{empty}
\pagestyle{empty}

\begin{abstract}
The paper focuses on the development of the autonomous robot UltraBot to reduce COVID-19 transmission and other harmful bacteria and viruses. The motivation behind the research is to develop such a robot that is capable of performing disinfection tasks without the use of harmful sprays and chemicals that can leave residues, require airing the room afterward for a long time, and can cause the corrosion of the metal structures. UltraBot technology has the potential to offer the most optimal autonomous disinfection performance along with taking care of people, keeping them from getting under UV-C radiation. The paper highlights UltraBot's mechanical and electrical structures as well as low-level and high-level control systems.
The conducted experiments demonstrate the effectiveness of the robot localization module and optimal trajectories for UV-C disinfection. The results of UV-C disinfection performance revealed a decrease of the total bacterial count (TBC) by 94\% on the distance of 2.8 meters from the robot after 10 minutes of UV-C irradiation.
\end{abstract}

\section{Introduction}
In the face of the COVID-19 world-girdling pandemic, it has become more apparent how important the disinfection of premises is to our lives. The spread of bacteria and viruses on surfaces and in the air is as dangerous as contact with an infected person. Researchers have shown that the most effective protection methods from COVID-19 are using masks and disinfection of surfaces \cite{COVIDMasksResearch}, \cite{COVIDHospitalResearch}. Applying masks helps only to prevent the spreading of some diseases, whereas the disinfection of air and surfaces is always effective. 

There are some ways for sanitizing premises. The most cost-effective way is manually cleaning surfaces where viruses spread, such as elevator buttons, door handles, and handrails. However, this method is very unreliable and time-consuming. The chances that a cleaner misses some of the infected areas rise as the area to be wiped increases. That is why this cleaning method is poorly suited for warehouses, shopping centers, and other places. Also, exposure to some of the disinfectants  can be dangerous for staff \cite{DangerousDesinfected}. 

Autonomous robotic systems have been gaining popularity in numerous fields of industry, and various tasks from stocktaking in warehouses \cite{kalinov2019high}, \cite{kalinov2020warevision} and plant disease detection \cite{karpyshev2021autonomous} to cleaning and sanitizing large areas. Different types of robots can clean the floor and clear recycle bins, disinfecting public spaces \cite{floor_can}. Autonomy gives two key advantages. First, robots can use more powerful disinfectants that pose a danger to humans. Secondly, their effectiveness does not decrease as the area that they need to sanitize increases. Such robots can potentially substitute humans, who are working within conditions that represent a threat to their health. 

\begin{figure}
\centering
\includegraphics[width=0.45\textwidth]{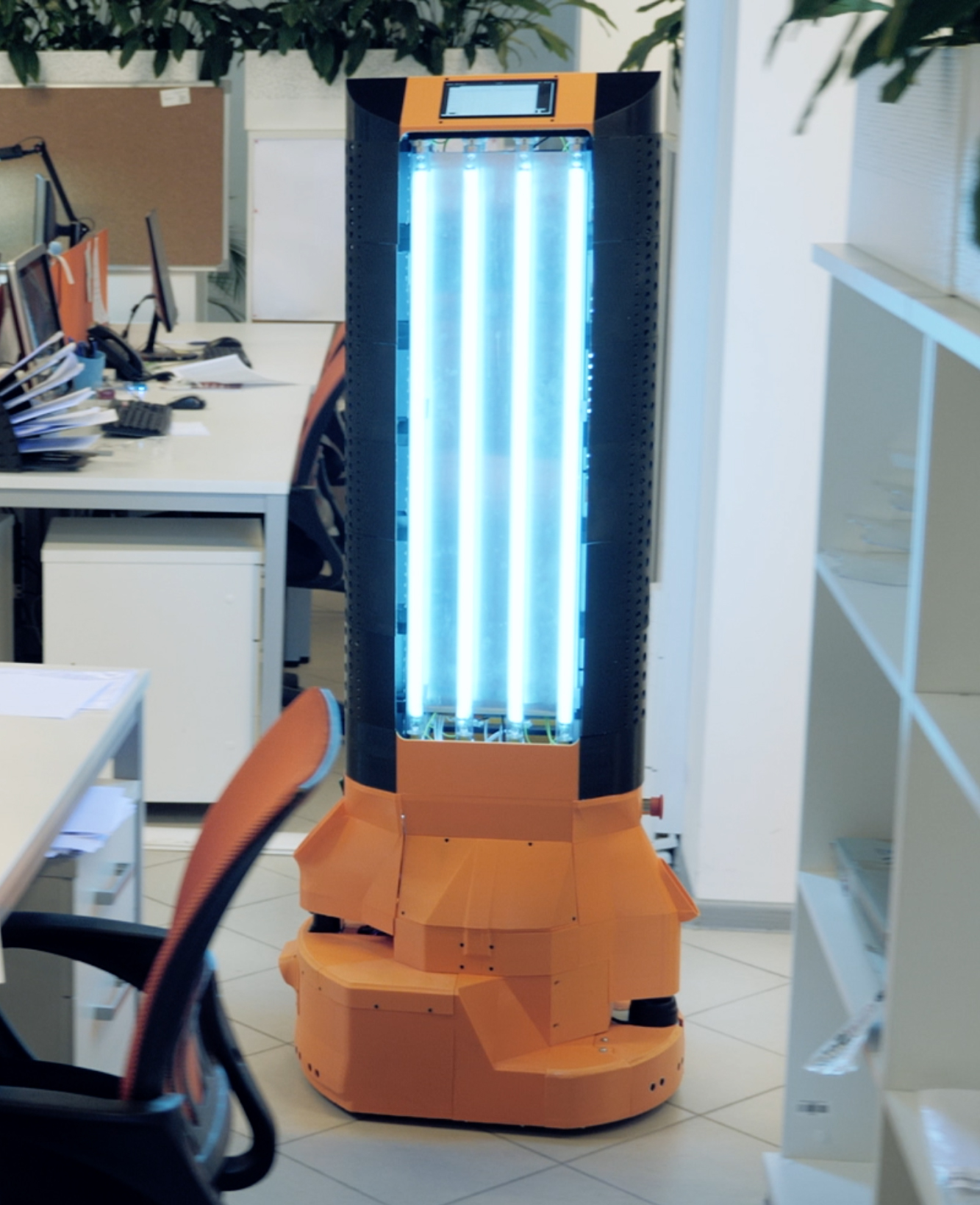}
\caption{UltraBot disinfecting company office.}
\label{UB}
\end{figure}

\subsection{Related works}
Current technical solutions for disinfection air and surfaces can be divided into two separate groups by some features. The first feature is the mobility of devices.
Non-mobile disinfecting systems can be represented by stationary installations with UV-C (ultraviolet C light) lamps, UV-based air flow cleaner-recirculator, and etc. A stationary germicidal box with UV-C lamps can clean only air in a small area near the device. However, the research showed that while sneezing large droplets (diameter between 60 \textmu m and 100 \textmu m) are carried away for more than 6 meters of horizontal distance \cite{TransmissionCOVID}. These droplets are too heavy to remain in the air, therefore, they fall on the floor or nearby surfaces. 
The disadvantage of mobile disinfecting robots is in the difficulty of navigating in dynamic, cluttered and narrow environment, and at the same time to avoid obstacles. However, this problem can be solved by DL (Deep Learning), SLAM technologies, and unique robot design. Some cleaning robots with DL already learn from humans. For example, a human instructor uses demonstrations to teach a robot how to perform different tasks such as passing narrow aisles or even cleaning a table \cite{icup}. Reconfigurable cleaning robot can be used to penetrate into the narrowest places without losing the speed of cleaning. A Tetris-inspired robot hTetro \cite{htetro} can perform this task using its specific navigation and reconfiguration algorithms \cite{htetro_nav}. 
A possible collision with a human in a narrow aisle is a challenging problem for the robot as well. However, some robots solve the issue by adapting their size and shape to the current situation that appeared during the work \cite{panthera}. On the one side, such mechanisms make it easier for the navigation module to work. On the other hand, they should navigate very robustly and be ready to withstand the strong impacts of the real world, which is often not achievable.

The second feature is the disinfectant used. The most common disinfection methods are spraying and UV-C light irradiation. Several variants of robots have been developed to use aerial disinfectant sprays. One of the simplest options is quadcopters for irrigating farms. In some countries, water in them was replaced with a sanitizer, thus they could clean the streets. One of the main developers in this area is the Chinese agriculture technology company XAG\footnote{https://www.xa.com/en}. Another solution is the Atlantis security robot\footnote{https://www.atlantis-h2020.eu/}, which has been fitted with a disinfectant sprinkler. 

Aerosol-type disinfecting robots are more effective than manual cleaning. However, there are several drawbacks that limit their applications. The main disadvantage of spraying is incomplete surface coverage. In this way, high quality of cleaning cannot be achieved. When it comes to indoor applications, long air ventilation is required to clean chemically contaminated areas. Importantly, the aerosol spraying and chemical compounds can cause corrosion of metal structures that are widely used in warehouses and shopping centers.  

Another popular disinfection method is applying UV-C lamps. For this task, UV-C light with a wavelength from 220 to 300 nm is used. During the pandemic, researchers have confirmed the effectiveness of the use of UV-C lamps for disinfection \cite{UVResearch}, \cite{UV222nmResearch}. There are several mobile robots that use UV-C light for sanitizing. Such robots have been developed by TMiRob\footnote{http://www.tmirob.com/}, UVD Robots\footnote{https://www.uvd-robots.com/}, MIT\footnote{https://www.csail.mit.edu/}, and other companies\footnote{https://www.avarobotics.com/}. 
They clean the air and all surfaces in the room that can be irradiated. The main disadvantage of their application is the danger to humans. UV-C irradiation can cause severe skin burns and eye injuries (photokeratitis). 

The basic design of a simple UV-C disinfection robot and study of UV-C irradiation influence on S. aureus is presented in \cite{RobotUVCIEEE}. 
Nevertheless, up until now, there is no research on the autonomous robot for UV-C disinfection.

In this work, we present UltraBot (see Fig. \ref{UB}), an autonomous mobile robot with UV-C lamps for indoor disinfection, which can work in a group of people without posing a threat to them and maintaining the effectiveness of the disinfecting process. 

\subsection{Problem statement}

The robot is designed to work in large areas, possibly with people. These can be warehouses, shopping centres, offices.  To accomplish this task, the following requirements were formulated:
\begin{itemize}
    \item \textit{Control system}: The robot has to have manual and autonomous mode.
    \item \textit{Area square}: The robot has to sanitize premises within 4 hours continuously.
    \item \textit{Human-aware movement and disinfection}: The robot has to avoid people and turn off its lamps if they could irradiate people.
\end{itemize}

The rest part of this article is structured as follows. Section II presents mechanical design of the autonomous robot. Electrical configuration of the robot is reported in Section III. Section IV focuses on a high-level robot control system. Section V introduces experimental results of disinfection by the robot, and Section VI concludes the paper.

\section{Mechanical Structure}

\subsection{Design of UltraBot}

We placed strong attention on the safety of neighboring humans. In contrast to typical UV-C lamp placement in a cylindrical shape, we split the lamps into two opposite sides of the robot, thus the UV-C emitting area would be restricted to only 180 degrees. Overall, it allows the robot to operate concurrently side by side with humans without downtime. In advance to specific lamp placing, there were several distinctive features applied that differ the robot from already existing market solutions.

The robot's chassis is made from a steel plate with attached aluminum profiles, which carry the payload. We have used 3D-printers with PLA plastic filament to produce the body shell of the robot.  The body shell itself was assembled with cells, fastened around the carcass. Cell structure with 3D-printed elements allowed to considerably decrease the cost of the robot comparing with industrial design. The robot's wheeled platform is driven by two motor-wheels and supplemented with the idle swivel trolley wheel, set for stabilization of the whole platform.  The electronics module is packed inside by installation on the sliding trays, which are handy for maintenance.

\begin{figure}[h]
\centering
\includegraphics[width=0.4\textwidth]{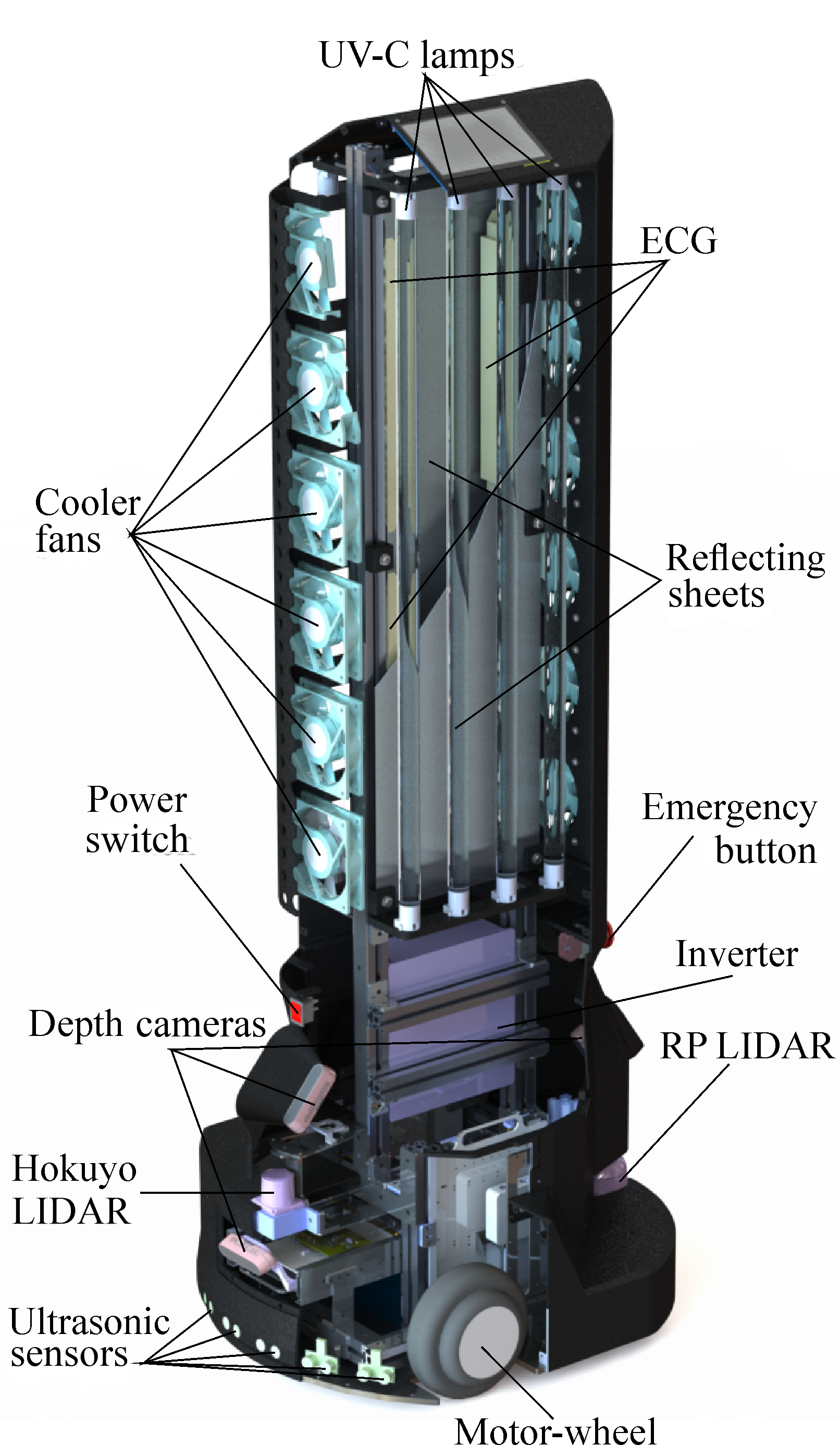}\label{true}
\caption{3D model of mobile robot for UV-C  disinfection.}
\label{fig:isometric}

\end{figure}

\subsection{UV-C radiation reflection}

As a result of the lamp stack's placement facing each other, the lamps evidently will not be able to work at full power because of the angle of glowing. To avoid the loss of disinfection power, the lamp module is supplied with a reflective shield made of celled anodized aluminium. This material is used in delousing chambers and has excellent performance. It has one of the highest retro-reflective indices for the full spectrum of ultraviolet rays. Among other things, it has high durability and corrosion-resistant properties. 

\subsection{Module for airborne particles disinfection }

In order to disinfect the air and add the ``safe mode'' of work, i.e. when the UV-C bulbs are fully covered by the shield in the future version, the robot was supplemented by an array of 12 fans. The fans are located lengthwise the lamps, thus the significant airflow volume passes through the maximum possible illuminating area. They are intended to work as the UV sterilizer in the situations when the robot is passing the humans in the safe mode. During this mode of work, the robot blows airstream through the disinfection system, functioning alike with a quartz lamp, simply maintaining airflow with the ionized particles and disinfecting the air.

\section{Electrical Structure}

\subsection {UV-C irradiation control}
As a disinfection technique, UV-C mercury vapor lamps were used. This choice is based on the effectiveness of such an approach. The UV-C fluorescent lamps are working at 253.7 nm wavelength, which has proven germicidal effectiveness \cite{UVResearch}-\cite{UVCGerm}.

Phillips TUV series lamps were chosen for this project. Each lamp has 12W of UV-C irradiation and 30W of power consumption. 
Lamps are divided into two isolated sides in a robot that could be driven separately. Four lamps on each side allow controlling of irradiation power. The control is carried out using relays that commute AC voltage to ECG drivers.

\subsection{Power system}



To power UV-C lamps, the 220 VAC inverter and ECG controllers being used to ignite the glow discharge between the cathode and anode of a lamp.
For the control system and additional electronics, DC-DC converters were used. The motor controllers are powered directly from the battery.   
The robot platform is facilitated with LiNCA battery pack with an overall capacity of 1200 Wh. It allows the robot to work autonomously for at least 4 hours with a single working side.

\subsection{Sensors}
As a primary sensor for robot navigation and localization, Hokuyo UST-10LX and RP2 LIDARs are used. The Hokuyo LIDAR is used as the main sensor for localization and mapping because of precision and update frequency. It is located in front of the robot's body. The second LIDAR determines a rear collision together with the main to achieve the field of view of 360 degrees. 

Ten ultrasonic sensors and four Intel RealSense cameras detect objects to avoid collision. Ultrasonic sensors are located under LIDARs to estimate small obstacles, stairs, and transparent objects. Cameras are used to detect the collision of a robot's upper part and people. Human detection is needed to emergency disable lamps if somebody stays near the working robot and for disinfection scheduling.


\begin{figure}[t]
\centering
\includegraphics[width=0.49\textwidth]{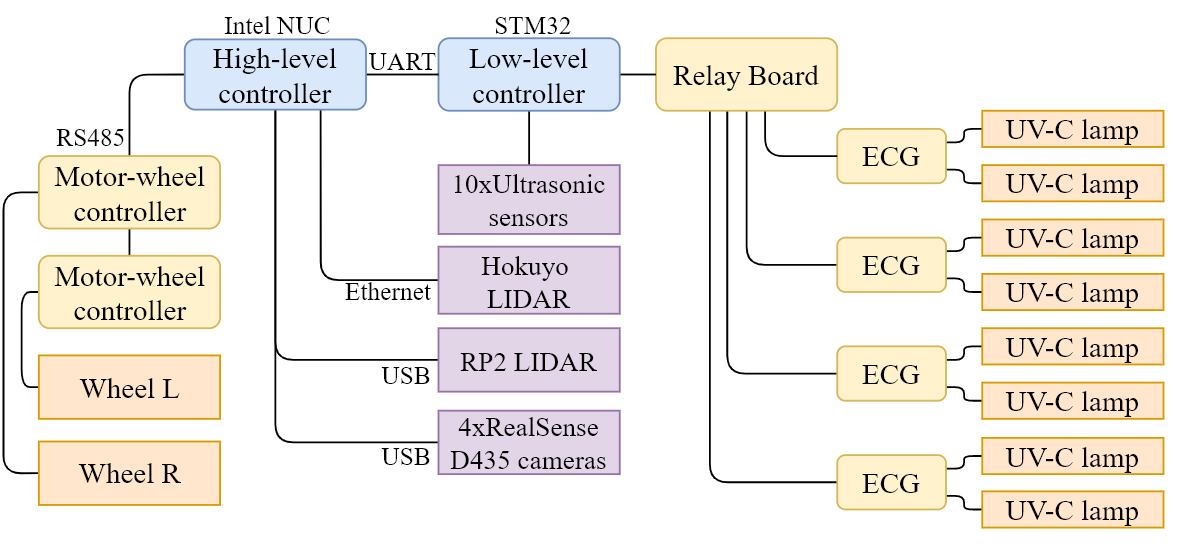}
\caption{Functional block diagram of the UltraBot control system.}
\label{fig:control}
\vspace{-1em}
\end{figure}

\subsection{Low-level control system}
The robot's control system is based on the high-level controller, i.e. Intel NUC computer, and low-level controller board. The low-level controller board based on STM32 microcontroller provides interfaces for ultrasonic sensors, UV-C lamp control, LED control, and battery status. The communication of high and low-level controllers is implemented with the UART interface. High-level Intel NUC controls LIDARs, cameras, and motor-wheels directly.

\section{High-level Control System}
As the robot employs dangerous for humans UV-C lamps, it must be fully autonomous to operate without any additional human support. Thus, the highly stable high-level control system was developed utilizing localization, mapping, navigation, low-level collision avoidance, PID control, and STM-control modules.

In order to make our system more convenient for users, we have considered 3 robot operational modes:
\begin{itemize}
    \item \textit{Manual control mode}: This option allows to control the robot movement by joystick. It is very convenient in case of the robot relocation instead of moving it by hands.
    \item \textit{Autonomous mode in an unknown environment}: This option assumes that the robot is placed in an unknown environment, where it should operate autonomously. The robot performs continuous simultaneous localization and mapping (SLAM) while planning and following the trajectories, to obtain knowledge of the environmental obstacles and avoid them.
    \item \textit{Autonomous mode in a known environment}: This option assumes that the robot already has a general map of the target environment and should operate autonomously there, taking into account dynamic obstacles.
\end{itemize}

\subsection{Localization}
In case of \textit{Autonomous mode in a known environment}, as a pure localization module we use \textbf{amcl} ROS package, a probabilistic localization system for a mobile robot moving in 2D \cite{amcl}. It implements the adaptive Monte Carlo localization approach, which uses a particle filter to track the pose of a robot against a known map.
However, in the case of \textit{Autonomous mode in an unknown environment}, we use Google Cartographer to perform the localization. It provides real-time SLAM in 2D \cite{google_cartographer}. In both cases, data from both forward and backward LIDARs is used to compute the robot's position.


\subsection{Mapping}
In our high-level control system mapping is performed by Google Cartographer. In the case of \textit{Autonomous mode in an unknown environment}, it performs SLAM, including localization work mentioned above.
In the case of \textit{Autonomous mode in a known environment}, this module's work is performed by \textbf{map server} ROS package that uploads already existing map to the operational process of high-level control system.


\subsection{Low-level collision avoidance}
A low-level collision avoidance algorithm is developed in order not to allow the robot to run into objects that appeared on its way \cite{decathlon}. The target module is active in all previously described operational modes. Thus, even in \textit{Manual control mode}, the robot doesn't collide with any obstacle. Low-level collision avoidance doesn't depend on a high-level control (navigation or localization modules). It makes motor-wheels ignore any high-level commands and reduces the wheel velocities down to 0 in case if an obstacle is on the robot's way. For the low-level collision avoidance algorithm, preventing possible collisions, data from both LIDARs and ultrasonic sensors are used.

\subsection{Navigation}
\begin{figure}[h]
\vspace{0.1em}
\centering
\includegraphics[width=0.45\textwidth]{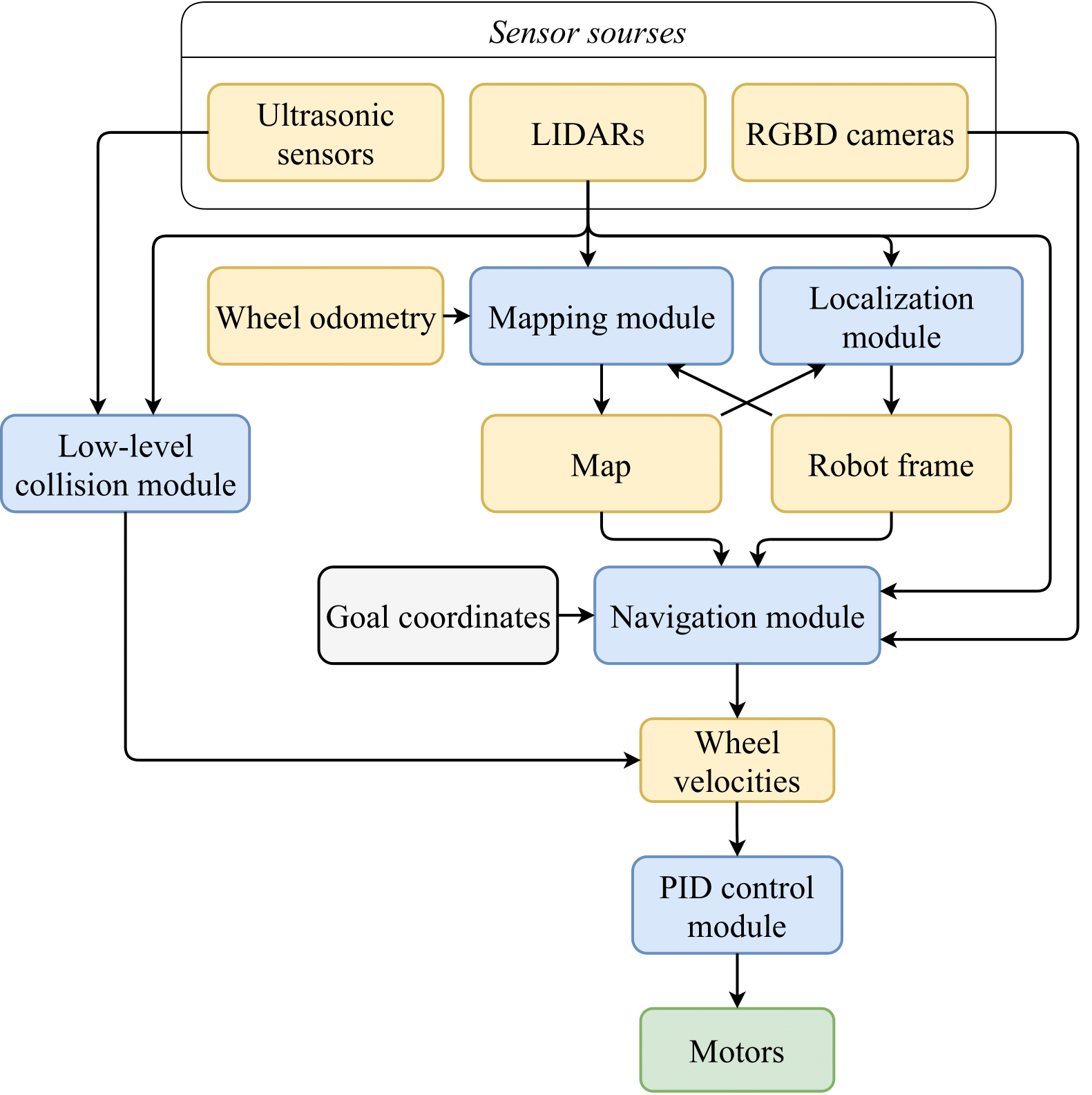}\label{true}
\caption{Navigation stack for robot \textit{Autonomous mode.}}
\vspace{-1em}
\label{fig:auto_unknown}
\end{figure}

The navigation module is based on \textbf{move base} ROS package, which inputs are sensor (LIDAR, RGB-D camera) data, localization data (robot coordinates), goal coordinates, and a map (continuously building or already existed). The output of the module is wheel velocities, which are sent to the PID control module and afterward to wheel motors. A navigation stack of robot \textit{Autonomous modes} is provided in Fig. \ref{fig:auto_unknown}. The map, robot frame, and goal coordinates are used by the navigation module to define the robot position on the map and build the trajectory from the robot to the goal point. LIDAR and RGB-D camera data is used to make the robot react to suddenly appeared dynamic obstacles and avoid them by rebuilding the robot trajectory.

\subsection{Control of UV-C lamps and LED}
In order to allow the high-level control system to manage UV-C lamps, the STM control module was developed. In case if a human is detected near the robot by RGB-D cameras, a command is immediately sent to STM board to deactivate the lamps. Besides this option, the STM control module is responsible for sending commands that modify the LED color. This allows to visually detect the following robot statuses:

\begin{itemize}
    \item Waiting for high-level commands.
    \item Moving to a goal.
    \item Avoiding an obstacle.
    \item A system error occurred.
\end{itemize}

This visual representation allows humans to quickly identify the robot's current status and helps in both trouble-shooting and making humans behave correctly while being close to the robot.

\section{Experimental Results}
\subsection{Navigation and localization}

One of the most important parameters that characterizes the stable operation of a mobile robot is the accuracy of localization. It's well known that robot localization is highly dependent on its navigation algorithms. Even a small, quick robot turn can make it lose its true position in the environment.

In order to determine the influence of navigation algorithms on localization work, several tests with different types of robot trajectories covering the space to be disinfected have been performed. We compared trajectories obtained by the robot itself with ground-truth data obtained by tracking the robot externally. For tracking the ground truth trajectory of the robot, we used the motion capture system Vicon Vantage V with 12 IR cameras covering a 27 ${m^3}$ space.

\begin{figure*}[!t]
\centering
\includegraphics[width=0.99\textwidth]{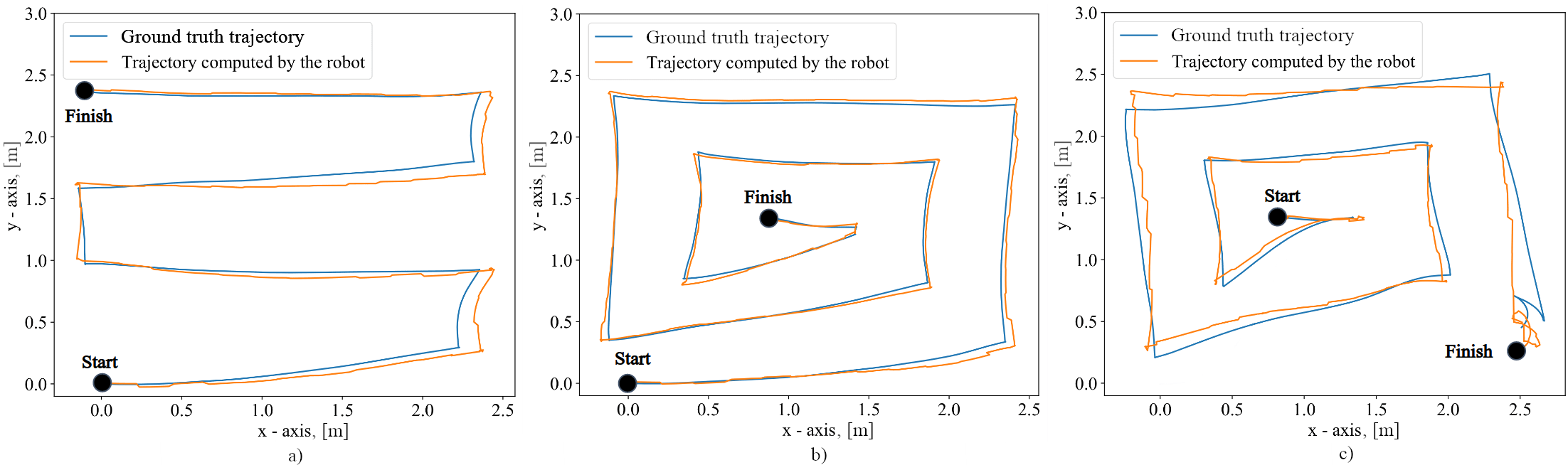}
\caption{Robot trajectories: a) S-shape. b) Rolling-up RPS. c) Unfolding RPS. }
\label{fig:trajectories}
\end{figure*}

\begin{figure}[!t]
\centering
\includegraphics[width=0.49\textwidth]{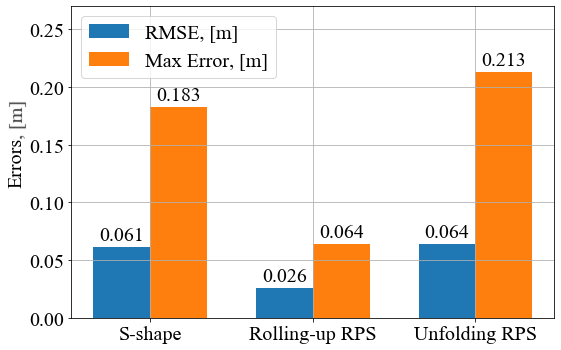}
\caption{Localization errors.}
\label{fig:errors}
\end{figure}

Three types of robot trajectories were considered, such as S-shape trajectory, Rolling-up Rectangular Planar Spiral (RPS) trajectory, and Unfolding RPS trajectory. All computed robot trajectories compared with ground-truth data are presented in Fig. \ref{fig:trajectories}. Resultant errors of the robot trajectories are provided in Fig. \ref{fig:errors}.

According to the obtained results, the lowest RMSE and Max error values correspond to the Rolling-up RPS trajectory when the robot moved to the room center. At the beginning of the trajectory, the robot moved very close to the obstacles (walls) which allowed its localization system to get a high level of confidence about the robot's position. Thus, the robot followed it during the whole work, even while moving away from the walls. What is more, this kind of trajectories is used by less complicated robots with a less number of sensors, for example, IDust robot, which navigation task starts from defining a position of room walls \cite{idust}.

The maximum error was achieved for the Unfolding RPS trajectory. Being far away from the obstacles which could allow the robot to localize, along with making a lot of turns, robot experiences difficulties in positioning. Comparing to the last one, the S-shape trajectory performed better, as soon as all robot turns occurred near the room walls.

This conclusion about preferable robot movement behaviour could be considered into setting up of a collaborative work of multiple disinfection robots, where generated robot trajectories must not intersect along with performing coverage path planning task \cite{multi_cleaning}.






\subsection{UV-C disinfection performance validation}

In order to validate the performance of the mobile robot in terms of disinfection by UV-C lamps, laboratory research has been conducted. For the research, several samples were taken at an industrial warehouse before and after the disinfection, which lasted up to 10 minutes. The samples were taken from cardboard boxes placed at different distances from the robot, and at different altitudes. Using the samples, laboratory research on the total bacterial count has been conducted in the Center of hygiene and epidemiology in St. Petersburg, Russia. The research results are listed in Table \ref{tab:disinfection}.

\begin{table}[h]
\vspace{1em}
    \centering
    \caption{UV-C Experimental Results}
    \begin{tabular}{|c|c|c|c|c|c|c|}
    \hline
    \multirow{2}{*}{N} & Distance & Height & \multicolumn{2}{|c|}{Total bacterial count} & TBC decrease\\
    \cline{4-5}
     & m & m & Before & After & \% \\
    \hline
    1 & 1.0 & 0.0 & 280 & 20 & 93 \\
    \hline
    2 & 2.8 & 0.0 & 180 & 10 & 94 \\
    \hline
    3 & 2.8 & 1.3 & 530 & 80 & 85 \\
    \hline
    4 & 5.0 & 0.0 & 230 & 130 & 43 \\
    \hline
    5 & 5.0 & 1.3 & 620 & 120 & 81 \\
    \hline
    6 & 10.0 & 0.0 & 220 & 70 & 68 \\
    \hline
    7 & 10.0 & 1.3 & 190 & 80 & 58 \\
    \hline
    \end{tabular}
    \label{tab:disinfection}
\end{table}

According to the results obtained, a total bacterial count (TBC) decreased to 93\% at the distance of 1 meter from the robot. On longer distances, an average TBC decrease was 90\%, 62\%, and 63\% on 2.8, 5.0, and 10.0 meters, respectively. Surprisingly, there is a significant decrease of the UV-C effectiveness between the ranges of 2.8 and 5.0 meters. In this area, the TBC decrease is highly dependent on the range, which is not true for other ranges. This leads to the conclusion that the limit of the UV-C range for the most effective disinfection (more than 90\%) is 2.8 meters.

\section{Conclusions}

We have presented the autonomous UV-C disinfection robot for indoor application with the possible presence of human beings. A new design of the robot with two separate UV-C lamp blocks has been developed. A high-level control system has been proposed and validated using the motion capture system Vicon Vantage V.

During experimental study we have identified the optimal disinfection distance for a robot, which should be less than 2.8 meters for a long disinfection time. The results of UV-C disinfection performance revealed a decrease of TBC by 94\% at the distance of 2.8 meters from the robot after 10 minutes of UV-C irradiation. Additionally, experimental results revealed that the optimal type of disinfection robot trajectory is the Rolling-up Rectangular Planar Spiral trajectory. This type of trajectory has the minimum RMSE of 0.026 meters in comparison with other curves. 

In future research, we will study the optimal parameters of robotic disinfection, i.e., speed of the robot and distance to the target surface. It is planned to find a non-trivial form of the robot disinfection zone for optimal path planning, which occurs due to two separate UV-C lamp blocks.
One of the main tasks to be solved is to predict human behavior for operative replanning of the robot trajectory without exposing people to UV-C irradiation. Therefore, human detection algorithm should be implemented. It will mark the points in place with high human activity and plan a disinfection route alongside them.

\bibliographystyle{IEEEtran}
\bibliography{ref}
\end{document}